\pdfoutput=1
\documentclass[11pt]{article}

\usepackage[final]{EMNLP2022}

\usepackage{times}
\usepackage{latexsym}
\usepackage[T1]{fontenc}
\usepackage[utf8]{inputenc}
\usepackage{microtype}
\usepackage{inconsolata}
\usepackage{hyperref}       % hyperlinks
\usepackage{url}            % simple URL typesetting
\usepackage{booktabs}       % professional-quality tables
\usepackage{amsfonts}       % blackboard math symbols
\usepackage{nicefrac}       % compact symbols for 1/2, etc.
\usepackage{microtype}      % microtypography
\usepackage{xcolor}         % colors
\usepackage{multicol}
\usepackage{graphicx}
\usepackage{caption}
\usepackage{subcaption}
\usepackage{pdfpages}
\usepackage{wrapfig}
\usepackage{enumitem}
\usepackage{multirow}
\usepackage{mfirstuc}
\usepackage{hhline}

\usepackage{epsfig}
\usepackage{graphicx}
\usepackage{float}
\usepackage{wrapfig}
\usepackage{algorithm,algorithmicx,algpseudocode}
\usepackage{bm,xspace}
\usepackage{comment}
\usepackage{verbatim}
\usepackage{multirow}
\usepackage{array}
\usepackage{balance}
\usepackage{url}
\usepackage{booktabs}
\usepackage{etoolbox,siunitx}
\usepackage{calc}
\usepackage{pifont,hologo}
 
\usepackage{nicefrac}

\usepackage{arydshln}
\usepackage[utf8]{inputenc}
\usepackage[T1]{fontenc}
\usepackage{url} 
\usepackage{amsfonts}
\usepackage{nicefrac}
\usepackage{microtype}
\usepackage[american]{babel}
\usepackage{enumitem}
\usepackage{amssymb}
\usepackage{amsmath}
\usepackage{siunitx}

\usepackage[super]{nth}
\usepackage{nicefrac}
\robustify\bfseries

\usepackage{dsfont}

\newcommand{\sD}{\mathcal{D}}

\newcommand{\sI}{\mathcal{I}}

\DeclareMathSymbol{@}{\mathord}{letters}{"3B}

\makeatletter
\DeclareRobustCommand\onedot{\futurelet\@let@token\@onedot}
\def\@onedot{\ifx\@let@token.\else.\null\fi\xspace}

\def\eg{\emph{e.g}\onedot} 
\def\ie{\emph{i.e}\onedot}

\makeatother

\definecolor{mypurple}{RGB}{104, 52, 154}
\definecolor{myblue}{RGB}{47, 110, 186}
\definecolor{myred}{RGB}{234, 51, 35}
\definecolor{myyellow}{RGB}{255, 194, 8}
\definecolor{mygreen}{RGB}{84, 130, 53}

\newcommand{\vl}{VL\xspace}
\newcommand{\vlendtoend}{VLE2E\xspace}
\newcommand{\ns}{NS\xspace}

\newcommand{\dsetshift}{Cross-Benchmark Transfer}
\newcommand{\compgen}{Compositional Generalization}
\newcommand{\contrast}{Contrast Set}
\newcommand{\segcomb}{Segment-Combine Test}

\newcommand{\segcombfirst}{segmentation\xspace}
\newcommand{\segcombsecond}{combination\xspace}

\newcommand{\testdsetshift}{\MakeLowercase{\dsetshift}\xspace}
\newcommand{\testcompgen}{\MakeLowercase{\compgen}\xspace}
\newcommand{\testcontrast}{\MakeLowercase{\contrast}\xspace}
\newcommand{\testsegcomb}{\MakeLowercase{\segcomb}\xspace}

\newcommand{\testcontrasts}{\MakeLowercase{\contrast{s}}\xspace}
\newcommand{\testsegcombs}{\MakeLowercase{\segcomb{s}}\xspace}

\newcommand{\ctestdsetshift}{\dsetshift\xspace}
\newcommand{\ctestcompgen}{\compgen\xspace}
\newcommand{\ctestcontrast}{\contrast\xspace}
\newcommand{\ctestsegcomb}{\segcomb\xspace}

\newcommand{\gqa}{GQA\xspace}
\newcommand{\vqa}{VQA\xspace}
\newcommand{\nlvrt}{NLVR2\xspace}
\newcommand{\nlvr}{NLVR\xspace}
\newcommand{\imsitu}{imSitu\xspace}
\newcommand{\covr}{COVR\xspace}
\newcommand{\clevr}{CLEVR\xspace}
\newcommand{\wino}{Winoground\xspace}
\newcommand{\traindata}{\sD_\text{train}}
\newcommand{\testdata}{\sD_\text{test}}

\newcommand{\visb}{VisualBERT\xspace}
\newcommand{\vilb}{ViLBERT\xspace}
\newcommand{\lxmt}{LXMERT\xspace}
\newcommand{\vinvl}{VinVL\xspace}

\newcommand{\tfive}{T5\xspace}
\newcommand{\bart}{BART\xspace}
\newcommand{\gptt}{GPT-2\xspace}

\newcommand{\largenewlogicform}{compositional logical forms\xspace}
\newcommand{\largeoriglogicform}{original logical forms\xspace}
\newcommand{\newlogicform}{CLF\xspace}
\newcommand{\origlogicform}{OLF\xspace}

\newcommand{\genexec}{\textsc{GenExec}\xspace}
\newcommand{\gtexec}{\textsc{GTExec}\xspace}
\newcommand{\exactm}{\textsc{Exact}\xspace}

\newcommand{\countgb}{\textsc{CountGroupBy}\xspace}
\newcommand{\vcountgb}{\textsc{VerifyCountGroupBy}\xspace}
\newcommand{\vcountgbdash}{\textsc{VerifyCount-GroupBy}\xspace}
\newcommand{\vcount}{\textsc{VerifyCount}\xspace}
\newcommand{\quant}{\textsc{Quantifier}\xspace}
\newcommand{\vquantattr}{\textsc{VerifyQuantAttr}\xspace}
\newcommand{\sameattr}{\textsc{SpecificSameAttr}\xspace}

\newcommand*\textnlp[1]{\textit{#1}}
\newcommand*\textlogic[1]{\texttt{#1}}

\title{Generalization Differences between End-to-End and Neuro-Symbolic Vision-Language Reasoning Systems}

\author{
\textbf{Wang Zhu}
\quad\quad \textbf{Jesse Thomason}
\quad\quad \textbf{Robin Jia} \\
University of Southern California, Los Angeles, CA, USA \\
\texttt{\{wangzhu, jessetho, robinjia\}@usc.edu}
}

\begin{document}
\maketitle
\begin{abstract}
For vision-and-language (\vl) reasoning tasks, both fully connectionist, end-to-end methods and hybrid, neuro-symbolic methods have achieved high in-distribution performance.
In which out-of-distribution settings does each paradigm excel?
We investigate this question on both single-image and multi-image visual question-answering through four types of generalization tests: a novel \testsegcomb for multi-image queries, \testcontrast, \testcompgen, and \testdsetshift.
Vision-and-language end-to-end (\vlendtoend) trained systems exhibit sizeable performance drops across all these tests.
Neuro-symbolic (\ns) methods suffer even more on \testdsetshift from \gqa to \vqa, but they show smaller accuracy drops on the other generalization tests and their performance quickly improves by few-shot training.
Overall, our results demonstrate the complementary benefits of these two paradigms, and emphasize the importance of using a diverse suite of generalization tests to fully characterize model robustness to distribution shift.
\end{abstract}

\begin{figure}
    \centering
    \includegraphics[width=\linewidth]{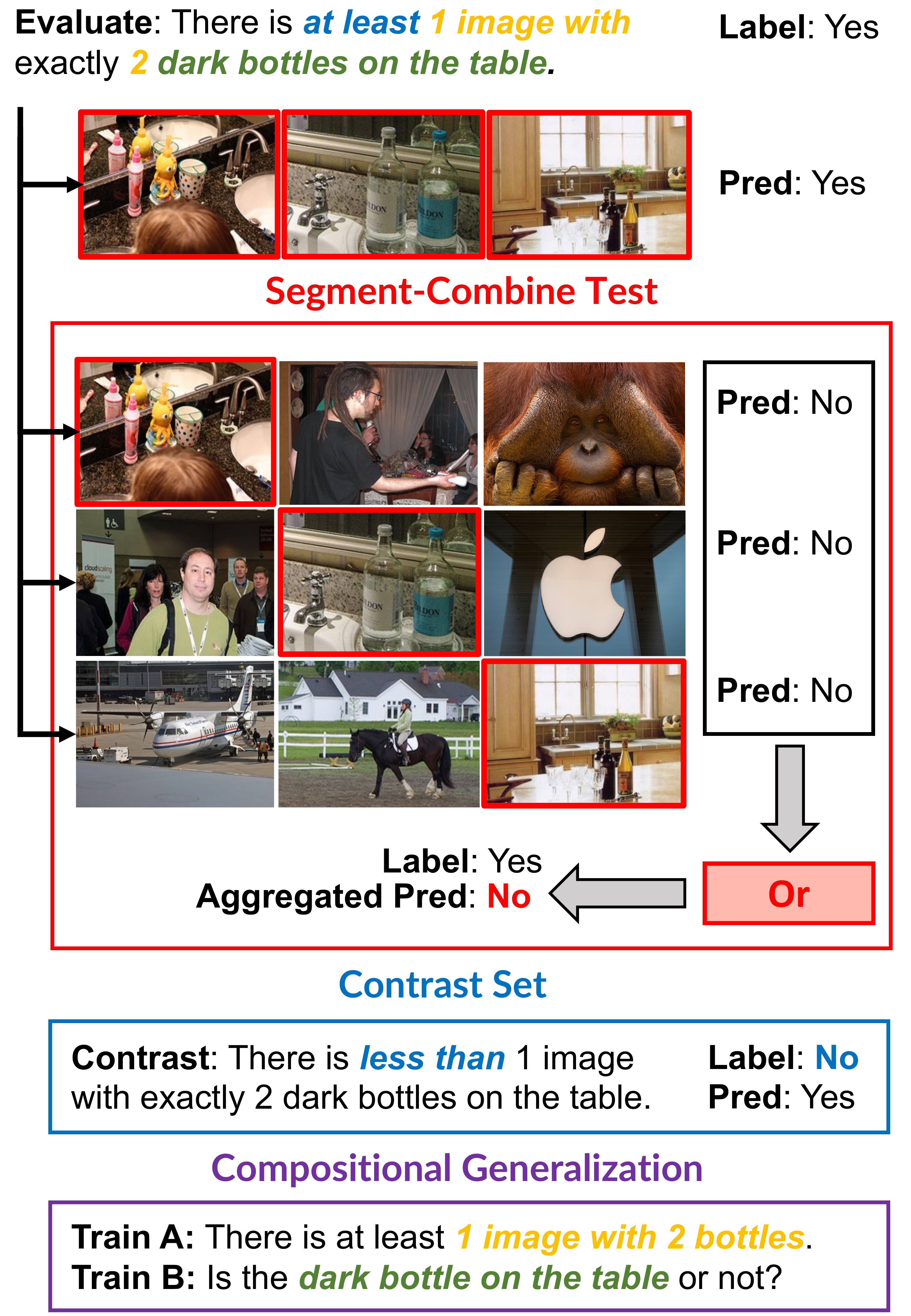}
    \caption{We build {\color{myred} \testsegcombs}, {\color{myblue} \testcontrasts} and {\color{mypurple} \testcompgen} splits for multi-image question answering in the COVR dataset. The above question requires counting both within and across images. {\bf \ctestsegcomb}: the multi-image query enables considering each image in isolation, pairing them with random unrelated images and feeding to the model, doing an OR operation of per-image answers.
    {\bf \ctestcontrast}: language perturbation by replacing phrases in query with synonyms or antonyms. {\bf \ctestcompgen}: the evaluated query is a compositional variant of questions Train A and Train B, involving reasoning on both {\color{myyellow} counting} and {\color{mygreen} relations}.}
    \label{fig:teaser}
\end{figure}

\section{Introduction}
\label{sIntro}
Widely used multi-modal pretrained models~\cite{chen2020uniter,lu2019vilbert,li2019visualbert} have exhibited great performance when fine-tuned on downstream vision-and-language tasks like \vqa~\cite{VQA} and \gqa~\cite{hudson2019gqa}. 
These models often generalize poorly to out-of-distribution (OOD) data, suggesting shortcomings in the \vlendtoend pipeline.
Neuro-symbolic methods~\cite{nsd, nsvqa} try to address this issue by disentangling grounding and reasoning mechanisms in multi-modal systems.
NS methods generate grounded visual representations, parse the language into executable programs for reasoning, and execute the programs on the visual representations.
Previous work~\cite{hudson2018compositional, Mao2019NeuroSymbolic} has shown the effectiveness of neuro-symbolic methods for OOD compositional generalization on single-image \vl reasoning tasks.
However, we still lack a comprehensive understanding of the generalization differences between these two paradigms under various setups.
Given recent work suggesting that OOD accuracy often strongly correlates with in-distribution accuracy \citep{miller2020effect,miller2021accuracy},
we might expect \vlendtoend and \ns systems to often have similar generalization abilities.
But do they?

In this work, we conduct the first comprehensive comparison of generalization behavior between \vlendtoend and \ns systems for \vl reasoning tasks.
Our study spans single-image and multi-image settings with natural images and includes four distinct types of generalization tests, three of which are shown in Figure~\ref{fig:teaser}.
We introduce a novel \textbf{\testsegcomb} for multi-image settings that requires models to make consistent predictions when some input images are replaced with irrelevant ones.
We evaluate on \textbf{\testcontrasts}~\citep{gardner-etal-2020-evaluating}, including new contrast sets we construct for \covr that test understanding of quantifiers.
We also measure \textbf{\testcompgen} as defined by compositional splits from \covr~\cite{bogin2021covr}
and \textbf{\testdsetshift} between \vqa and \gqa.
We also develop improved \ns systems for \gqa by handling mismatches between program and scene graph object descriptors, and for \covr by refining the original logical language.

Overall, we find that \vlendtoend and \ns systems exhibit distinct and complementary generalization patterns. 
The \ns systems are more robust than the \vlendtoend systems in the first three testing situations.
The \vlendtoend systems exhibit overstability to meaning-altering perturbations, suggesting they overfit to spurious correlations in the training data and do not learn precise reasoning skills.
We further find that the semantic parsing module of \ns systems can quickly improve on generalization tests given a few training examples, whereas \vl models do not adapt as quickly.
On the other hand, while \vlendtoend systems lose more than 10\% in accuracy on transfer between \vqa and \gqa, the \ns methods perform even worse.
Taken together, our findings underscore the need for a diverse suite of generalization tests to fully compare different modeling paradigms.
The different behavior of these two systems could guide the community to design more robust \vl reasoning systems.
We release our code for generating test data, and we encourage future \vl models to be evaluated on these tests.\footnote{We release our code and test data at \url{https://github.com/Bill1235813/gendiff\_vlsys}}
\section{Related Work}
\label{sRelated}

We first survey related work on vision-language reasoning models and OOD evaluation tests. 

\paragraph{\vl OOD Generalization.} 
Many efforts have been made to evaluate the generalization ability of \vlendtoend systems and task-specific methods on compositionality~\cite{johnson2016clevr, Thrush_2022_CVPR}, language perturbations~\cite{ribeiro-etal-2019-red} and visual perturbations~\cite{jimenez-etal-2022-carets}.
\citet{linjie2020closer} showed \vlendtoend systems exhibit better robustness than task-specific methods. 
We are the first to comprehensively compare the generalization differences between \vlendtoend and \ns systems across different OOD tests.

\paragraph{\vl Pretrained Models.}
Large-scale, \vl pretrained models for question-answering can be single-stream---encoding vision and language features together with a single transformer---such as \visb~\cite{li2019visualbert} and \vinvl~\cite{Zhang_2021_CVPR}, or dual-stream---encoding vision and language with separate transformers and applying cross-modal transformers later---such as \vilb~\cite{lu2019vilbert} and \lxmt~\cite{tan2019lxmert}.
We evaluate on both single- and dual-stream \vl pretrained models.

\paragraph{Neuro-Symbolic Methods.}
NS-VQA~\cite{nsd} disentangled vision and language processing for \vl reasoning tasks on simulated images.
However, it requires the datasets to include annotations of logical forms to describe language.
To reduce the supervision signal from program annotations, NS-CL~\cite{Mao2019NeuroSymbolic} jointly learned concept embeddings and latent programs, and extended to natural images.
NSM~\cite{hudson2019learning} learned graph-level reasoning and showcased the compositional reasoning abilities of \ns methods.
To be applicable to both single- and multi-image setups, we choose the same pipeline as in the original NS-VQA.
We use the scene graph as the structural representation, and test on multiple language models for semantic parsing.

\paragraph{Single- and Multi-Image \vl Reasoning Tasks.}
For \vl reasoning, there are many datasets that focus on single images,
such as \clevr~\cite{johnson2016clevr}, \vqa, and \gqa,
as well as many other datasets that involve multi-image reasoning, such as \nlvr~\cite{suhr-etal-2017-corpus}, \nlvrt~\cite{suhr2019corpus}, \covr~\cite{bogin2021covr}, and \wino~\cite{tristan2022wino}.
We experiment with two single-image datasets, \vqa and \gqa, and one multi-image dataset, \covr, all of which use natural images.

\section{Models}
\label{sProblem}
Next, we formally define the \vl reasoning tasks and \vlendtoend and \ns methods we study.
We also discuss a new \ns system for \covr and associated changes to the original \covr logical forms.

\subsection{Vision-Language Reasoning}
In a \vl reasoning task, each example consists of a triple $(q, \sI, y)$, where $q$ is a natural language query, $\sI$ is a set of queried images and $y$ is the corresponding answer of the query. 
The number of queried images is $|\sI|$, \eg, $|\sI|=1$ for a single-image query.
Given query $q$ and image set $\sI$, a VL system $f$ predicts an answer $\hat{y}=f(q, \sI)$.
Models are trained on $\traindata$ and evaluated on $\testdata$. 

\begin{figure}[t]
    \centering
    \includegraphics[width=\linewidth]{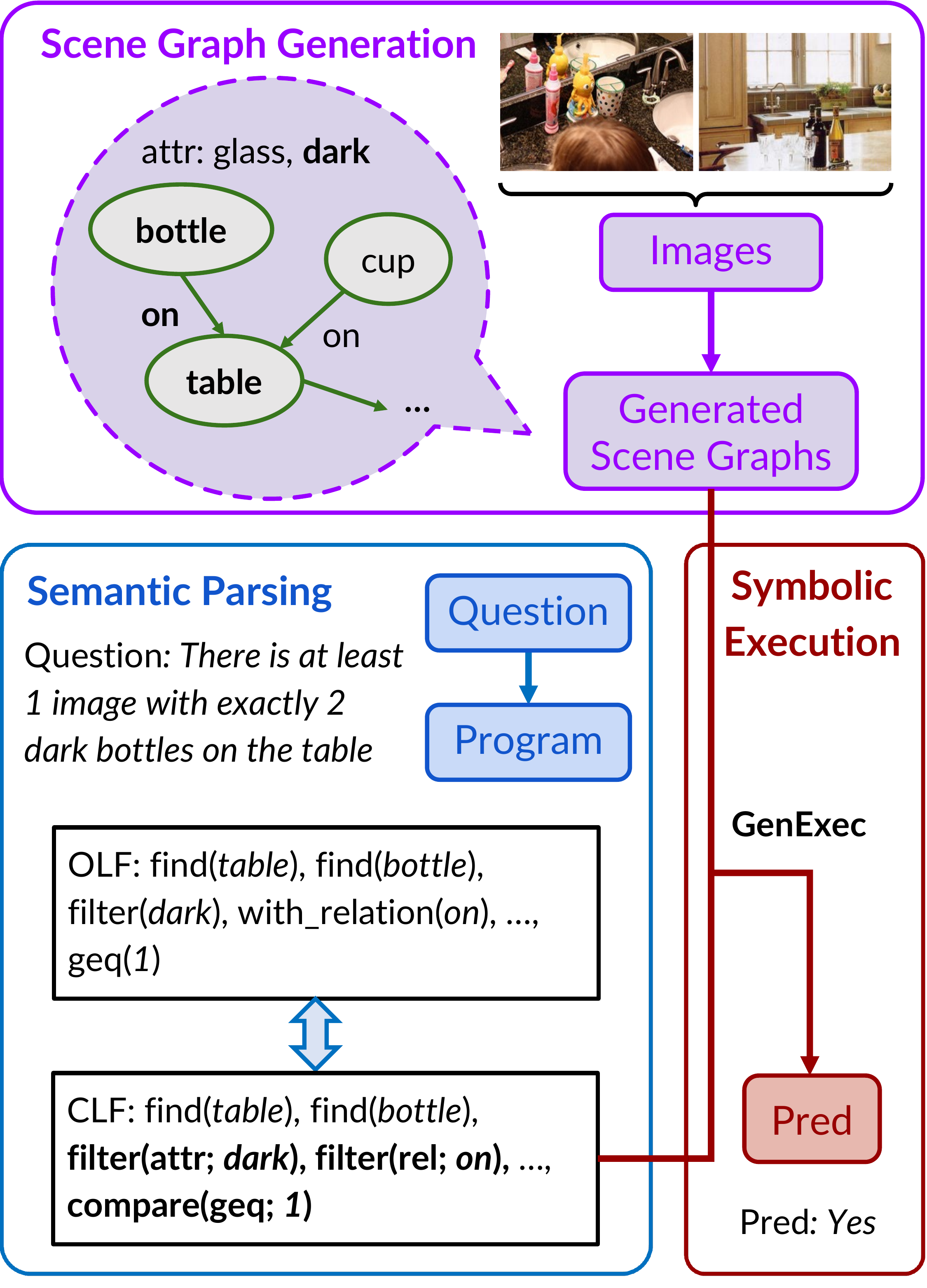}
    \caption{The process of the multi-image query with the modified neuro-symbolic methods. A language model ({\color{myblue} blue}) maps the question to a functional program in our compositional logical forms (CLF) format; differences with the original logical forms (OLF) are shown in \textbf{bold}. A scene graph generator ({\color{mypurple} purple}) processes each image into a separate scene graph; queried information shown in \textbf{bold}. The program is executed on all the scene graphs together to produce an answer ({\color{red} red}).}
    \label{fig:ns_model}
\end{figure}

\subsection{Modified \vlendtoend System}
For a \vlendtoend system, $f$ is a single neural network that is trained end-to-end.
Since current \vl pretrained models are trained to process single images,
we modify the \vlendtoend pipeline for multi-image settings following~\citet{bogin2021covr}.
Given a multi-image query $(q, \sI)$ and a pretrained model, for each image $I \in \sI$, we feed the pair $(q, I)$ to the pretrained model to get an image-text representation. 
We concatenate these $|\sI|$ image-text representations and prepend a \texttt{[CLS]} token to construct a sequence of length $|\sI|+1$. 
We then input this generated sequence into a two-layer transformer, and take the produced embedding of the \texttt{[CLS]} token as the representation of the entire multi-image query.
Finally, we feed the representation into an MLP classifier to predict $y$.
All modules including the pretrained model are fine-tuned.
We experiment with 4 different \vl pretrained models: the single-stream \visb and \vinvl and the dual-stream \lxmt and \vilb.
    
\subsection{Modified \ns System}
A \ns system separately processes vision and language with two trainable modules $\phi$ and $\psi$. 
The image set is represented as $\phi(\sI)$, and the query semantics is represented as a functional program $\psi(q)$.
A pre-defined executor executes $\psi(q)$ on $\phi(\sI)$ to predict the answer $\hat{y}$.
To apply NS-VQA-like pipelines to real-world images, we use scene graphs as the structured representation $\phi(\sI)$. 

We use a pre-trained scene graph generator that can be fine-tuned on task-specific scene graph data, depending on the dataset (see \S\ref{sec:exp-setup} for details).
We fine-tune large language models to map queries $q$ to functional programs $\psi(q)$ (i.e., semantic parsing).
We experiment with 3 language models: (1)~\tfive~\cite{raffel20t5}, (2)~\bart~\cite{lewis-etal-2020-bart} and (3)~\gptt~\cite{radford2019language}. 

Now, we describe dataset-specific work needed to build a full \ns pipeline for \gqa and \covr. 
Both datasets provide logical forms for each question, but these forms require modification to be compatible with \ns systems. 

\paragraph{Single-Image Queries.} 
In \gqa, functional programs align with objects in scene graphs via object IDs. 
For example, a program may refer to object \textnlp{``bird(775)''}, while the corresponding node for object 775 in the scene graph could have the name \textlogic{parrot}. 
Since object IDs are not predictable by a semantic parsing model given $q$, we remove them from the annotated programs. 
Thus, we need to ground object references like \textnlp{``bird''} to likely coreferents like the \textlogic{parrot} node.
We construct a dictionary that maps each object type mentioned in a program in $\traindata$ (e.g., \textnlp{``bird''}) to the set of all scene graph object types that such a mention matched to (e.g., \textlogic{parrot}).
We use this dictionary to match objects between programs and scene graphs when executing programs at test time.
Mismatches between object names in the program and scene graph occur in 9.5\% of validation examples, but using this dictionary resolves 99.6\% of the mismatches.

\paragraph{Multi-Image Queries.} 
Like \gqa, multi-image queries in \covr are annotated with executable programs and ground truth scene graphs of images. 
The program annotation incorporates quantifier operations, which enables \ns execution of the multi-image queries without changing the pipeline of the NS-VQA methods~\cite{nsvqa}. Figure~\ref{fig:ns_model} provides an overview of our multi-image NS pipeline.

In the compositional splits for \covr, models must generalize to some unseen compounds (\eg, phrases) consisting of seen tokens (\eg, words). For example, models could be tested on \textnlp{``Is the child sitting on a branch or a swing?''} after seeing \textnlp{``What is the child sitting on?''}, \textnlp{``Is the child sitting on a swing?''} and \textnlp{``Is the child sitting on a branch?''} at training time.
However, the annotated logical forms in \covr for the above test query include an unseen unit operation \textlogic{choose\_name} (used to choose either ``branch'' or ``swing''), which is not possible to generate as it was not seen at training time.
To at least make compositional generalization possible, we design a set of compositional logical forms as an intermediate representation~\cite{herzig2021unlocking} based on the existing programs in \covr.
For the operation \textlogic{choose\_name}(\textnlp{branch, swing}), we take the prefix ``choose'' as the operation name and leave the postfix ``name'' as an argument, the new operation is \textlogic{choose}(\textlogic{name}, \textnlp{branch, swing}). By doing so, it becomes possible to generate this operation once we see a \textlogic{choose}(\textlogic{attr}, $\cdot$) and a \textlogic{query}(\textlogic{name}, $\cdot$) operation.
We try to keep a minimum set of operations by redesigning non-composable operations and eliminating redundant operators. 
By doing so, we reduce the size of the operation set from 33 to 17. 
We cover the details of the modified programs in Appendix~\ref{suppClf}. 
We denote the new logical forms as \largenewlogicform (\newlogicform) in contrast to the \largeoriglogicform (\origlogicform), and evaluate the \ns system based on these programs for the generalization tests.

\paragraph{Evaluation Metrics.} 
We use 3 different evaluation metrics for the \ns system. 
Our main evaluation metric is \genexec, the accuracy with the program execution on the generated scene graphs.
To measure the effect of errors during scene graph generation errors, we also measure \gtexec, the accuracy with the program execution on the ground truth scene graphs.
Finally, we also measure \exactm,
the exact match accuracy of the programs generated by semantic parsing;
this penalizes ``spuriously correct'' parses that execute to the right answer but compute the wrong function.

\begin{table*}[ht]
    \centering
    \small
    \tabcolsep 4pt
    % \resizebox{1\linewidth}{!}{
    \begin{tabular}{llccccc}
        \toprule
        Template & Transfer Setting & \multicolumn{2}{c}{\vlendtoend} & \multicolumn{3}{c}{\ns} \\
        \cmidrule(lr){3-4}\cmidrule(lr){5-7}
        & & \visb & \vilb & \newlogicform w/ \bart & \newlogicform w/ \gptt & \newlogicform w/ \tfive \\
        \midrule
        \multirow{2}{*}{\countgb} & Original Query & \bf 55.8 & 52.4 & 48.8 [99.0] & 49.0 [99.7] & 49.0 [99.9]\\
        &  Segment-Combine & 44.6 & 40.1 & 48.7 [98.9] & 48.9 [99.6] & {\bf 49.0} [99.7] \\
        \midrule
        \multirow{2}{*}{\vcountgb} & Original Query & 72.1 & \bf 73.6 & 70.7 [99.5] & 70.8 [99.6] & 70.9 [99.7] \\
        &  Segment-Combine & 52.5 & 55.7 & 70.7 [99.3] & {\bf 70.8} [99.6] & {\bf 70.8} [99.7] \\
        \bottomrule
    \end{tabular}
    % }
    \caption{\segcomb\xspace results on \covr. \vlendtoend models fail on both counting questions (\countgb) and binary questions (\vcountgb), while all \ns models are robust. Oracle \gtexec results for \ns models are in brackets.
    }
    \label{tab:seg_comb}
\end{table*}

\section{Evaluation Methods}
\label{sMethods}
We evaluate \vlendtoend and \ns systems on four generalization tests.
We create a new multi-image perturbation test called the \segcomb, and create new contrast sets for \covr by perturbing quantifiers. 
We also test models on compositional generalization and cross-benchmark transfer.

\paragraph{\segcomb.} 
We introduce the segment-combine test to test model generalization on multi-image perturbations. 
For a multi-image query $(q, \sI)$ where $\sI=(I_1, ..., I_{|\sI|})$, we first perform a \segcombfirst phase.
We make $|\sI|$ queries,
where the $k$-th query uses the original question $q$ and an image set formed by the union of the original image $I_k$ plus $|\sI|-1$ random images \textit{unrelated} to $q$. 
We feed these to the model to get $|\sI|$ predictions. 
Next, in the \segcombsecond phase, we apply an aggregation function (e.g., \textlogic{SUM} or \textlogic{OR}) based on the question type to fuse these predictions (Figure~\ref{fig:teaser}).
A robust model should return the same answer on the segment-combine test and the original example.

We run the \testsegcomb on \covr, sampling random images from all images in the \covr validation set. 
To confirm that we only sample images unrelated to the original image set (\ie, will not change the answer after fusion), we execute ground truth programs on ground truth scene graphs for each query in the \segcombfirst phase, and find that the accuracy is 100\%.

We focus on two templates in \covr for which there is an appropriate fusion function. For the template \countgb (\eg, \textit{``How many images have 2 bottles?''}), the fusion function is \textlogic{SUM}.
That is to say, the answer on the original input should be equal to the sum of the $|\sI|$ answers from the \segcombfirst phase. For the template \vcountgb, the fusion function is logical \textlogic{OR}, as shown in Figure~\ref{fig:teaser}.

\paragraph{\contrast s.}
For \vl reasoning, we define a contrast set~\cite{gardner-etal-2020-evaluating} of an example $(q, \sI, y)\in\testdata$ to be a set of similar examples $(q', \sI, y')$, where $q'$ is similar to $q$ and $y'$ may or may not be the same as $y$, depending on $q'$.
$q'$ could be constructed by replacing specific words or phrases in $q$ with synonyms or antonyms, or by substituting objects with other objects. 
Given $n$ contrast set examples $(q_1', \sI, y_1'), \dotsc, (q_n', \sI, y_n')$,
we primarily evaluate models on the average accuracy on these $n$ examples.
We also measure the average \textit{local coherency} as $\frac{1}{n}\sum_{i=1}^n (\hat{y_i}=\hat{y'_i})$, which measures how much the model ignores perturbations.

We use the single-image \testcontrasts created by \citet{contrastGQA} for \gqa.
Their \testcontrasts involve object substitutions from scene graphs and mainly test the robustness of \vlendtoend systems for grounding objects.

For multi-image \covr, we design new \testcontrasts that target perturbations involving cross-image reasoning for multi-image queries. 
We replace quantifiers in the testing data with phrases of the equivalent and opposite meanings and change the labels accordingly. 
We focus on examples generated by 4 templates, where quantifiers (\eg, \textnlp{at least, all}) play the role of introducing cross-image reasoning:
one counting question template \countgb, and three binary question templates, \vcountgb, \vcount (\eg, \textit{``At least 2 bottles on the table?''}) and \quant (\eg, \textit{``No bottles are on the table?''}).
We test meaning-preserving perturbations such as replacing \textnlp{at least} with \textnlp{no less than} on counting and binary questions.
We also test meaning-altering perturbations such as replacing \textnlp{no} with \textnlp{some} on binary questions and flipping the answer. 
We do not apply meaning-altering perturbations to counting questions as it is non-trivial to determine what the new answer $y'$ should be.

\paragraph{\ctestcompgen.} 
In this setting, $\traindata$ and $\testdata$ are from the same benchmark, but the queries in $\testdata$ are compositional variants of those in $\traindata$.
For example, $\testdata$ examples may contain two phrases that were seen independently in $\traindata$ but never together. We test on the compositional generalization splits as defined in \covr, which are constructed by holding out a question template or holding out the examples where multiple query properties co-occur during training.

\paragraph{\ctestdsetshift.} 
In this setting, $\traindata$ and $\testdata$ are from different benchmarks. 
We choose one of \vqa and \gqa as $\traindata$ and the other as $\testdata$.

\begin{table*}[ht]
    \centering
    \small
    \tabcolsep 12pt
    % \resizebox{1\linewidth}{!}{
    \begin{tabular}{lcccccc}
        \toprule
        Eval data & \multicolumn{3}{c}{\vlendtoend} & \multicolumn{3}{c}{\ns} \\
        \cmidrule(lr){2-4}\cmidrule(lr){5-7}
        & \lxmt & \vilb & \vinvl & w/ \bart &
        w/ \gptt & w/ \tfive \\
        \midrule
        \gqa-Val & 83.9 & 83.5 & \bf 89.1 & 65.6 [78.1] & 71.8 [80.4] & 74.3 [85.1] \\
        \gqa-Val-Contrast & 66.5 & 68.2 & 73.3 & 64.8 [78.8] &  71.9 [82.3] & {\bf 74.1} [85.0] \\
        \bottomrule
    \end{tabular}
    % }
    \caption{\contrast\xspace results on \gqa. \vlendtoend shows $\sim$15\% of performance drop even if the contrast set is an easy object substitution, while \ns models are highly robust. Note that \gqa-Val is a subset of the validation set of \gqa used to create the contrast set \gqa-Val-Contrast. Oracle \gtexec results for \ns models are in brackets.}
    \label{tab:single_os}
\end{table*}

\section{Experiments}
\label{sExps}

We present our experimental setup and results on four types of generalization tests below. The results indicate the complementary robustness of \vlendtoend and \ns systems in OOD settings.

\subsection{Experimental Setup}
\label{sec:exp-setup}
We use \vqa~\cite{VQA} and \gqa~\cite{hudson2019gqa} as our single-image QA dataset and \covr~\cite{bogin2021covr} as our multi-image QA dataset. \vqa has three types of questions: binary, (yes/no), counting (answer is a number) and open-ended (answer can be any term from a vocabulary).

For the \testdsetshift between \vqa and \gqa, as \vqa and \gqa have different sets of labels, we filter both validation sets to only include labels that appear in both datasets. Note that \vqa has no program and scene graph annotations, so we can only train the \ns methods on \gqa.

For model training, we use the fine-tuning setups described in the respective papers for each model.
We give further details about hyperparameter selection in Appendix B.
For \ns methods, we generate scene graphs with the unbiased scene graph generation method Causal-TDE~\cite{tang2020unbiased}, which uses Faster R-CNN~\cite{ren2015faster} as the backbone for object detection.

\begin{table*}[t]
    \centering
    \small
    \tabcolsep 2pt
    % \resizebox{1\linewidth}{!}{
    \begin{tabular}{p{2.5cm}p{3cm}ccccccc}
        \toprule
        Template & Transfer Setups & OOD & FL & \multicolumn{2}{c}{\vlendtoend} & \multicolumn{3}{c}{\ns} \\
        \cmidrule(lr){5-6}\cmidrule(lr){7-9}
        & & & & \visb & \vilb & \newlogicform w/ \bart & \newlogicform w/ \gptt & \newlogicform w/ \tfive \\
        \midrule
        \multirow{2}{*}{\countgb} & \textnlp{at least} $\rightarrow$ \textnlp{at least} & & & \bf 53.3 & 53.0 & 48.9 [99.0] & 49.0 [100.0] & 49.0 [100.0] \\
        & \textnlp{at least} $\rightarrow$ \textnlp{no less than} & \checkmark & & \bf 54.0 & 52.9 & 23.7 [43.6] & 21.7 \phantom{1}[38.8] & 21.5 \phantom{1}[39.4] \\
        \midrule
        \multirow{3}{*}{\vcount} & \textnlp{at least} $\rightarrow$ \textnlp{at least} & & & \bf 87.6 & 84.3 & 75.1 [99.1] & 75.6 [100.0] & 75.6 [100.0] \\
        & \textnlp{at least} $\rightarrow$ \textnlp{no less than} & \checkmark & & \bf 68.3 & 67.0 & 49.9 [69.8] & 48.2 \phantom{1}[67.4] & 48.4 \phantom{1}[67.8] \\
        & \textnlp{at least} $\rightarrow$ \textnlp{less than} & \checkmark & \checkmark & 21.1 & 24.5 & 52.3 [64.0] & {\bf 53.0} \phantom{1}[65.5] & 52.9 \phantom{1}[65.6] \\
        \midrule
        
        \multirow{3}{*}{\parbox[t]{2.5cm}{\vcountgbdash}} & \textnlp{at least} $\rightarrow$ \textnlp{at least} & & & \bf 74.8 & 70.5 & 71.4 [99.3] & 71.6 \phantom{1}[99.5] & 71.4 \phantom{1}[99.5] \\
        & \textnlp{at least} $\rightarrow$ \textnlp{no less than} & \checkmark & & \bf 65.8 & 63.2 & 56.2 [77.5] & 55.0 \phantom{1}[76.0] & 55.6 \phantom{1}[76.2] \\
        & \textnlp{at least} $\rightarrow$ \textnlp{less than} & \checkmark & \checkmark & 28.5 & 32.1 & 50.8 [51.8] & 49.9 \phantom{1}[50.4] & {\bf 51.1} \phantom{1}[50.6] \\
        \midrule
        \multirow{8}{*}{\quant} & \textnlp{no} $\rightarrow$ \textnlp{no}, \textnlp{some} $\rightarrow$ \textnlp{some} &  &  & 86.5 & \bf 90.2 & 75.2 [97.4] & 75.5 \phantom{1}[97.9] & 75.5 \phantom{1}[97.8] \\
        & \textnlp{no} $\leftrightarrow$ \textnlp{some} & \checkmark & \checkmark & 74.9 & \bf 80.6 & 75.1 [97.3] & 75.7 \phantom{1}[97.9] & 75.6 \phantom{1}[97.8] \\
        \cmidrule{2-9}
        & \textnlp{no} $\rightarrow$ \textnlp{no} & & & \bf 94.4 & 92.1 & 77.3 [96.7] & 77.6 \phantom{1}[97.0] & 77.4 \phantom{1}[96.9] \\
        & \textnlp{no} $\rightarrow$ \textnlp{at least one} & \checkmark & \checkmark & \bf 83.9 & 78.6 & 53.7 [61.2] & 54.5 \phantom{1}[61.4] & 53.9 \phantom{1}[61.1] \\
        \cmidrule{2-9}
        & \textnlp{some} $\rightarrow$ \textnlp{some} & & & 80.3 & \bf 88.7 & 75.4 [98.3] & 75.6 \phantom{1}[98.7] & 75.6 \phantom{1}[98.6] \\
        & \textnlp{some} $\rightarrow$ \textnlp{none of the} & \checkmark & \checkmark & 67.8 & 68.2 & 71.9 [92.1] & 75.3 \phantom{1}[97.3] & {\bf 75.5} \phantom{1}[97.8] \\
        \cmidrule{2-9}
        & \textnlp{all} $\rightarrow$ \textnlp{all} & & & \bf 80.2 & 79.9 & 77.0 [98.2] & 78.0 \phantom{1}[98.9] & 78.1 \phantom{1}[99.1] \\
        & \scriptsize{\it all} $\rightarrow$ \scriptsize{\it either none or only some} & \checkmark & \checkmark & 36.2 & 39.2 & 54.0 [66.4] & 55.1 \phantom{1}[67.6] & {\bf 57.5} \phantom{1}[69.8] \\
        \bottomrule
    \end{tabular}
    % }
    \caption{\contrast\xspace results on \covr. OOD: OOD test; FL: Flip labels. \vlendtoend has drastic performance drops on some of the meaning-altering perturbations, while \ns shows equally performance drops regardless of meaning changes. Oracle \gtexec results for the \ns models are in brackets.}
    \label{tab:contrast_set-covr-1}
\end{table*}
\begin{figure*}[ht]
    \centering
    \includegraphics[width=\linewidth]{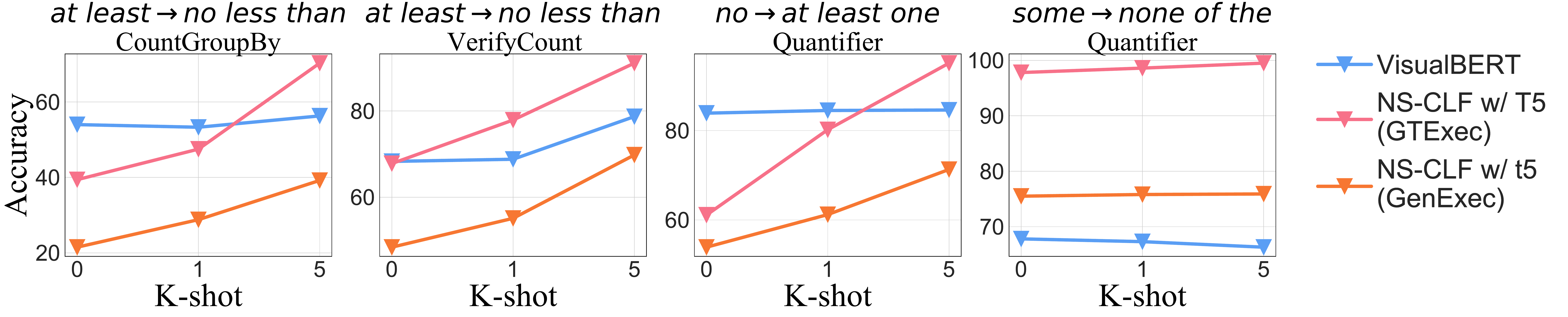}
    \caption{Few Shot Training on the \covr \contrast\xspace. The \ns model with \tfive as the semantic parsing module quickly improves performance with 5 new training examples. \visb does not improve as much.}
    \label{fig:few_shot}
\end{figure*}

\subsection{\segcomb}
\label{sec:result-segcomb}
We test \vlendtoend and \ns systems with \testsegcomb and list their accuracy in Table~\ref{tab:seg_comb}.

\paragraph{\vlendtoend models fail on the \testsegcomb.}
Both \visb and \vilb fail on the \testsegcomb, but \ns models achieve accuracy close to the original query. 
The performance drop of \vlendtoend models is 11-12\%  on counting questions (\vcountgb) and 18-20\% on binary questions (\countgb), as shown in Table~\ref{tab:seg_comb}. 
Though \ns models with generated scene graphs show 1-7\% lower accuracy than \vlendtoend models on the original multi-image queries, they achieve 4-18\% higher accuracy on the segment-combine evaluation data.

\paragraph{\vlendtoend models learn multi-image spurious correlations.}
We notice \visb's performance on the segment-combine test for binary questions (52.5\%) is close to random guessing.
Thus, we extract the prediction from \visb on the \testsegcomb.
For binary questions, 93\% of the prediction are \textnlp{no}.
For counting questions with 6 labels, 38\% of the predictions are \textnlp{0}. 
As \covr queries are created by sampling related and distracting images, \vlendtoend models tend to predict \textnlp{no} or \textnlp{0} for queries with more irrelevant images, which is a spurious correlation between queried images learned during fine-tuning.
By contrast, semantic parsing produces the right program to execute with \exactm score above 98.5\% for all \ns models, not just spuriously correct programs which are accidentally correct during execution.

\begin{table*}[ht]
    \centering
    \small
    \tabcolsep 5.5pt
    \begin{tabular}{lcccccc}
        \toprule
        Split & \multicolumn{1}{c}{Random Guess} & \multicolumn{2}{c}{In-distribution} & \multicolumn{3}{c}{Compositional Generalization}  \\
        \cmidrule(lr){2-2}\cmidrule(lr){3-4}\cmidrule(lr){5-7}
        & \multicolumn{1}{c}{Text only} & \multicolumn{1}{c}{\vlendtoend} & \multicolumn{1}{c}{\ns} & \multicolumn{1}{c}{\vlendtoend} & \multicolumn{2}{c}{\ns} \\
        \cmidrule(lr){3-3}\cmidrule(lr){4-4}\cmidrule(lr){5-5}\cmidrule(lr){6-7}
        & & \visb & \newlogicform w/ \tfive & \visb & \newlogicform w/ \tfive & \origlogicform w/ \tfive \\
        \midrule
        \textsc{TPL-ChooseObject} & 52.0 & 62.6 & 62.0 & \phantom{1}1.6 & {\bf 46.2} \phantom{1}[50.9] & \phantom{1}0.0 \phantom{12}[0.0] \\
        \textsc{TPL-VerifyQuantAttr} & 50.4 & 76.9 & 70.5 & \bf 71.2 & 44.8 \phantom{1}[48.4] & \phantom{1}0.0 \phantom{12}[0.0] \\
        \textsc{TPL-VerifyAttr} & 49.6 & 75.4 & 67.6 & \phantom{1}0.0 & {\bf 24.9} \phantom{1}[34.4] & \phantom{1}0.0 \phantom{12}[0.0] \\
        \midrule
        \textsc{Has-Count \& Has-Attr} & 41.2 & 62.6 & 72.4 & 58.7 & {\bf 70.9} \phantom{1}[99.1] & 68.1 \phantom{1}[93.4] \\
        \textsc{Has-Count \& RM/V/C} & 40.5 & 82.2 & 76.1 & 74.1 & {\bf 76.1} [100.0] & 75.0 \phantom{1}[98.4] \\
        \textsc{Has-SameAttr-Color} & 49.8 & 71.2 & 67.7 & 66.0 & {\bf 67.7} [100.0] & {\bf 67.7} [100.0] \\
        \bottomrule
    \end{tabular}
    \caption{\compgen\xspace on \covr. The upper panel shows splits with held-out templates. The lower panel shows splits with held-out property combinations. In-domain random guessing accuracy is from a \visb text only model. \newlogicform improves the accuracy of \ns on \testcompgen, and outperforms \visb on most of the tests.
    Oracle \gtexec results for the \ns models are in brackets.}
    \label{tab:init_cg}
\end{table*}

\subsection{\contrast s}
\label{sec:result-contrast}

We test on the augmented \gqa contrast set from~\citet{contrastGQA} for single-image queries, and compare the performance on the corresponding portion of \gqa validation data. 
We also test \vlendtoend and \ns systems on our generated contrast set on \covr involving cross-image reasoning. 

\paragraph{\vlendtoend models show weak object grounding.}
For perturbations that only involve object substitution, \lxmt, \vilb, and \vinvl show a performance drop of 15-17\%, as shown in Table~\ref{tab:single_os}.
This drop implies the \vlendtoend training is not robust even on object grounding. 
Although the \ns methods are worse than \vlendtoend systems on in-domain test data, they are highly robust on language-side object substitutions.
Our \ns pipeline with \tfive outperforms the best \vlendtoend method by 0.8 points on the contrast set, despite being 14.8 points worse on the in-domain test data.
This finding indicates the benefits on robustness of having a separate object grounding module.

\paragraph{\vlendtoend suffers on meaning-altering perturbations.}
For perturbations involving cross-image reasoning, both \visb and \vilb perform worse on meaning-altering perturbations than meaning-preserving perturbations, as shown in Table~\ref{tab:contrast_set-covr-1}.
For meaning-preserving perturbations, we observe no major accuracy drop on the counting questions, and a performance drop of 10-20\% on the binary questions.
On meaning-altering perturbations, replacing \textnlp{at least} and \textnlp{all} causes a more drastic performance drop of 40-65\% for both \visb and \vilb, while exchanging \textnlp{no} and \textnlp{some} only leads to 10\% drop. 
Our hypothesis is \vlendtoend systems cannot generalize well to logical operations that are rare in the fine-tuning data:
the opposites of \textnlp{at least} and \textnlp{all} rarely or never appear in the training data, whereas the opposites of \textnlp{no} and \textnlp{some} (i.e., \textnlp{some} and \textnlp{no}, respectively) are common.
The local coherency is 96.3\% for \textnlp{at least}$\rightarrow$\textnlp{less than} and 80.2\% for  \textnlp{all}$\rightarrow$\textnlp{either none or only some}, which implies the \vlendtoend systems do not pay enough attention to quantifiers whose opposites were not seen during fine-tuning.

\paragraph{\ns performance has no correlation with meaning change.}
The \ns methods, instead, show similar performance drop for both the meaning-preserving and the meaning-altering perturbations. 
The accuracy is higher than \vlendtoend models on most meaning-altering perturbations, but lower on the meaning-preserving ones, especially on counting questions.
In some meaning-altering cases, the oracle accuracy is even close to 100\%, which shows that the semantic parser is very robust in those situations.

\paragraph{\ns recovers quickly with few-shot training.}

We add 1 to 5 examples from a contrast set to the full training dataset and re-train the model for few-shot learning.
Figure~\ref{fig:few_shot} shows \ns methods learn quickly and adapt to the new example types, while \visb learns slowly under few-shot training. 
With gold scene graphs, the \ns accuracy increases even more quickly, suggesting that some improvements are hidden by the fact that our generated scene graphs are imperfect.
Note that for the \ns systems, we only adapt the language modeling part, as the contrast sets only affect language.
Thus, we can also conclude language-only models adapt faster than \vl neural models.

\subsection{Compositional Generalization}
\label{sec:result-compgen}
We test \vlendtoend and \ns systems with \covr \testcompgen sets and list their accuracy in Table~\ref{tab:seg_comb}. For the \ns systems, we compare our \largenewlogicform (\newlogicform) to the \largeoriglogicform (\origlogicform) from \covr.

\paragraph{\newlogicform improves generalization.}
Comparing the last two columns in Table~\ref{tab:init_cg}, it is clear that the new \newlogicform logical forms improve generalization to new combinations of query properties compared to original logical forms, and make generalization to new templates possible. 

\paragraph{\ns has lower in-domain but higher \testcompgen performance.}
In Table~\ref{tab:init_cg}, the in-domain accuracies of the \ns system are always lower than those of the \vlendtoend systems.
However, on most of the compositional splits, the performance of \visb is worse than the \ns method. 
The only exception is \vquantattr, where there is a complex operation comparing whether two lists of objects have some same attributes. 
Following our hypothesis of \vlendtoend is better at questions with phrases occurring frequently in training, we compute the cosine similarity of the text embedding in \vilb, and find examples in the template \vquantattr are semantically close to examples in the template \sameattr. Examples for templates \vquantattr and \sameattr are \textnlp{``Do all cats that are on a floor have the same color?''} and \textnlp{``Does the dog that is in grass and the dog that is in water have the same color?''}, respectively. However, these two templates have different logical forms in both \newlogicform and \origlogicform, making it easier for \vlendtoend systems to generalize but harder for \ns systems.

\subsection{Cross-Benchmark Transfer}
\label{sec:result-dsetshift}
The cross-benchmark test aims to explore transferability between benchmarks of the same visual question-answering task. We evaluate the transfer between \gqa and \vqa because they share similar types of queries.

\paragraph{\vlendtoend is more transferable than \ns.}
In Table~\ref{tab:gqavqa_cross}, \lxmt has an 8-15\% accuracy drop for transfer from each dataset to the other. 
However, the \ns method with \tfive as the semantic parser has even worse performance on transfer.
Using the scene graph generator and semantic parser trained on \gqa, the accuracy of the \ns method drops by over 70\% on open questions.

\paragraph{Failure of \ns is mainly due to scene graph generation error}
To understand the reasons for the failures of the \ns system, we conduct manual analysis on 40 \vqa examples. 
We observe that more than 75\% of the \vqa programs are correctly generated with the semantic parser trained on \gqa. 
However, they often do not execute to the correct answer because 
(1) semantically similar objects have different node names in the generated scene graphs; 
(2) some objects are harder to detect due to visual domain shift. 
For example, for a generated program like \textlogic{[``operation'': ``select'', ``argument'': ``mattress'']}, we may not find an object named ``mattress'' in the generated scene graph, where it could be named ``bed''.
To quantify this issue, we compute the missing object ratio, the percentage of programs that throw errors during execution because objects mentioned in the program are not found in the scene graph. 
The high missing object ratio in Table~\ref{tab:gqavqa_cross} suggests that the scene graph generation module trained on \gqa cannot correctly match objects mentioned in the programs for \vqa images.

\begin{table}[t]
    \centering
    \small
    \tabcolsep 7.5pt
    \begin{tabular}{llcc}
        \toprule
        Transfer Setups &  & \multicolumn{1}{c}{\vlendtoend} & \multicolumn{1}{c}{\ns} \\
        \cmidrule(lr){3-3}\cmidrule(lr){4-4}
        &  & \lxmt & w/ \tfive \\
        \midrule
         \multirow{2}{*}{VQA$\rightarrow$VQA} & Binary & 97.4 & - \\
         & Open & 90.0 & - \\
         \cmidrule(lr){2-4}
         \multirow{2}{*}{GQA$\rightarrow$VQA} & Binary & \bf 86.4 & 52.0 (78.4) \\
         & Open & \bf 78.7 & \phantom{1}8.3 (80.2) \\
         \cmidrule(lr){2-4}
         \multirow{2}{*}{GQA$\rightarrow$GQA} & Binary & \bf 90.6 & 78.6 \phantom{1}(3.8) \\
         & Open & \bf 81.8 & 60.1 \phantom{1}(5.3) \\
         \cmidrule(lr){2-4}
         \multirow{2}{*}{VQA$\rightarrow$GQA} & Binary & 82.5 & - \\
         & Open & 66.2 & - \\
        \bottomrule
    \end{tabular}
    \caption{The accuracy of \lxmt versus our \ns method trained on $X$ and deployed on the \textit{validation} set of $Y$ ($X\rightarrow Y$) for \vqa and \gqa.
    \ns methods are not able to train on \vqa due to lack of scene graph and program annotations (marked with dashes). \ns methods show bad transfer performance, especially on the open-ended questions, mainly due to the high ratio of programs that cannot find the queried objects from the generated scene graphs (marked in parentheses). \vlendtoend has less accuracy drop compared to \ns.}
    \label{tab:gqavqa_cross}
\end{table}

\paragraph{NS occasionally requires new primitives.}
Another possible reason for the \ns system's cross-benchmark failure from \gqa to \vqa would be that some question types in \vqa require new primitive operations.
In our manual analysis, less than 10\% of the \vqa programs require the addition of new primitive operations, demonstrating that this is not the primary reason for \ns struggles.
Most of these questions involve commonsense reasoning, such as asking why some event happens in the image (e.g., \textnlp{``Why is the man on the street?''} where the answer is \textnlp{``homeless''}). 
We also note that we only evaluate on the binary and open questions of \vqa but exclude the counting questions, which are about 13\% of the dataset. 
\gqa has no counting questions, so the semantic parser trained on \gqa cannot generate counting operations.

\paragraph{How much manual adaption is required to transfer \ns systems to a new benchmark?}
\ns systems require manual adaptation for different datasets. 
From the transfer between \gqa and \vqa, we show little manual adaption is required on the language side of \ns systems to transfer between benchmarks of the same task.
With some entity matching mechanism between semantically similar objects and a stronger scene graph generation module that generalizes well between datasets, \ns might be possible to transfer well.

\section{Discussion and Conclusion}

In conclusion, \vlendtoend training systems do not learn precise reasoning, which inhibits their generalization ability under small perturbations to either language or vision.
Though the in-domain results of \ns systems are usually slightly worse than \vlendtoend systems, the \ns methods are more robust on most of the generalization tests we develop here. 
Even when the performance of \ns methods drops on some OOD data, they can still quickly recover by few-shot training. 
Nonetheless, \vlendtoend systems still achieve better performance on \testdsetshift, while \ns methods struggle when test questions require novel program constructs or scene graph object types.

Our work highlights the importance of evaluating on a diverse set of metrics besides in-distribution accuracy, in line with recent work on improving leaderboards \citep{ethayarajh-jurafsky-2020-utility, ma2021dynaboard}.
Our analysis suggests that we should not expect in-domain and out-of-domain accuracies to be strongly correlated when evaluating very different types of models, such as \vlendtoend and \ns models, in contrast with \citet{miller2020effect, miller2021accuracy}. 
Finally, we hope our observation that end-to-end and neuro-symbolic systems have complementary generalization advantages will inspire the community to design more robust VL reasoning systems that share the benefits of both approaches. 

\section*{Acknowledgements}
This work was supported in part by the NSF (RI AWD-00001042, award number 1833137).

\section*{Limitations}
Most of our experiments focus on datasets with synthetic language annotations.
In particular, \gqa and \covr both use synthetic language, while \vqa has human-written questions.
Existing VL reasoning datasets with natural language questions do not have annotated functional programs and scene graphs. 
Since we use \ns systems must be trained on annotated programs,
we cannot easily extend our work to these other datasets.
One possible solution would be to adapt other single-image \ns methods (\eg, NSM~\cite{hudson2019learning}) that do not require program and scene graphs annotation to the multi-image setup.

Our evaluation requires a custom modification of the semantic parsing language on \gqa and \covr. 
To apply similar evaluations to other datasets, if their program annotations are not directly applicable to our \ns system, practitioners might need to make similar task-specific modifications.

Finally, all of our experiments are on English-only data, which requires limited morphological reasoning reasonable semantic parsing. 
The results and conclusions might not be applicable to other language with richer morphology.

% This document has been adapted by Yue Zhang, Ryan Cotterell and Lea Frermann from the style files used for earlier ACL and NAACL proceedings, including those for 
% ACL 2020 by Steven Bethard, Ryan Cotterell and Rui Yan,
% ACL 2019 by Douwe Kiela and Ivan Vuli\'{c},
% NAACL 2019 by Stephanie Lukin and Alla Roskovskaya, 
% ACL 2018 by Shay Cohen, Kevin Gimpel, and Wei Lu, 
% NAACL 2018 by Margaret Mitchell and Stephanie Lukin,
% Bib\TeX{} suggestions for (NA)ACL 2017/2018 from Jason Eisner,
% ACL 2017 by Dan Gildea and Min-Yen Kan, NAACL 2017 by Margaret Mitchell, 
% ACL 2012 by Maggie Li and Michael White, 
% ACL 2010 by Jing-Shin Chang and Philipp Koehn, 
% ACL 2008 by Johanna D. Moore, Simone Teufel, James Allan, and Sadaoki Furui, 
% ACL 2005 by Hwee Tou Ng and Kemal Oflazer, 
% ACL 2002 by Eugene Charniak and Dekang Lin, 
% and earlier ACL and EACL formats written by several people, including
% John Chen, Henry S. Thompson and Donald Walker.
% Additional elements were taken from the formatting instructions of the \emph{International Joint Conference on Artificial Intelligence} and the \emph{Conference on Computer Vision and Pattern Recognition}.

\bibliography{anthology}
\bibliographystyle{acl_natbib}

\clearpage
\appendix

\section{Compositional Logical Forms}
\label{suppClf}
We create \largenewlogicform as an intermediate representation of the original logical forms. We explain how to design new operations and default value grammar checker below.

\subsection{Operation Modifications}
\label{suppOpera}
As shown in Table~\ref{tab:clf_operations}, we refactor the quantifier operations to a \textlogic{map} operation with logical \textlogic{or} or logical \textlogic{and} as arguments.
To compositionally represent the original \textlogic{none} operation, we introduce a new operation \textlogic{logic\_not}, which takes a boolean variable and outputs the negation.
This enables the negation of the original operation \textlogic{all}, and also enables the contrast set creation from \textnlp{all} $\rightarrow$ \textnlp{either none or only some}.

We refactor and create \textlogic{choose}, \textlogic{query}, \textlogic{verify}, \textlogic{filter}, \textlogic{keeep\_if\_values\_count} and \textlogic{compare} operations following the same pattern.
We merge the redundant operation \textlogic{relation\_between\_nouns} with the \textlogic{choose} operation and replace two sanity check operations as \textlogic{unique} and \textlogic{assert\_unique} with an automatic grammar checker.
For the rest of operations, we keep them in \newlogicform as the original version.

\subsection{Default Value Grammar Checker}
\label{suppGram}
The default value grammar checker takes the sanity check responsibility from the original \textlogic{unique} operation. However, unlike the original \textlogic{unique}, which raises an error when the queried object is not unique, the grammar checker automatically fixes the error by taking the first object as the queried object. 
For example, for a query \textnlp{``Is the boy wearing a hat?''}, if the \textlogic{find} operation returns multiple \textnlp{``boy''} nodes, the grammar checker automatically choose the first \textnlp{``boy''} node and records an error. 
Otherwise, if the \textlogic{find} operation returns zero \textnlp{``boy''} nodes, the grammar checker will raise an ``object not found'' error, same as the missing object error in Table~\ref{tab:gqavqa_cross}.

To make sure the program is executable, the default value grammar checker also assigns a default value for the arguments for each operation. 
The default value is \textnlp{0} for the integer type and \textnlp{False} for the boolean type.
For example, if a \textlogic{compare} operation has only one argument, the value will directly compare to \textnlp{0}.

\section{Experiment Details}
\label{suppExps}

\subsection{Dataset Statistics}
\label{suppsubDsetStat}

The image data of \covr consists of \gqa and \imsitu~\cite{yatskar2016}. 
There are about 275k images in total. 
\vqa has human annotated queries, and the language annotation of \gqa and \covr is generated by template.
\vqa has 440k training questions, 214k validation questions and 448k testing questions. 
\gqa has 943k training questions, 132k validation questions and 95k testing questions. 
For \covr, each example contains 1 to 5 query images, and there are 248k training questions, 7k validation questions and 7k testing questions.

For datasets with scene graph annotations, \gqa and \covr, before generating scene graphs for the images in each dataset's validation set, both the object detection and the relation prediction modules in the scene graph generator are tuned on the training set.
For datasets without scene graph annotations, \vqa, we directly apply the scene graph generator trained on \gqa to generate scene graphs, and apply the constructed dictionary on \gqa to map objects.

\begin{table}[ht]
    \centering
    \resizebox{1\linewidth}{!}{
    \begin{tabular}{ccccc}
        \toprule
        Model & \# Params & \multicolumn{3}{c}{Running time per experiment} \\
        \cmidrule(lr){3-5}
        & & \covr & \gqa & \vqa \\
        \midrule
        \tfive & 220M & 15 & 15 & - \\
        \gptt & 117M & 15 & 15 & - \\
        \bart & 110M & 16 & 16 & - \\
        \midrule 
        \vinvl & - & - & 0 & - \\
        \vilb & - & 17 & 0 & - \\
        \visb & - & 17 & 0 & - \\
        \lxmt & - & - & 4 & 8 \\
        \bottomrule
    \end{tabular}
    }
    \caption{GPU hours are computed at 1 Quadro RTX 6000 GPU. \textnlp{0} indicates the model is downloaded without further training.}
    \label{tab:model_params}
\end{table}
\subsection{Hyperparameters}
\label{suppsubHyper}
The model parameters and GPU hours are listed in Table~\ref{tab:model_params}. 
We search hyperparameters manually, one per trial. 
For all language models, we choose a batch size of 256, train 30000 iterations at maximum, with an early-stopping with patience 3. 
We save the model each 500 iterations.
For \tfive, we use an Adam optimizer with a learning rate 1e-4. For \gptt, The learning rate is 5e-5 for \gptt and 2e-5 for \bart. 

For all vision models on \covr, we choose a batch size of 12, train 8 epochs with no early-stopping and save the best model from the evaluation after each epoch.
We use the AdamW optimizer with a learning rate of 1e-6 and a weight decay of 1e-3 for \visb and \vilb. \visb and \vilb take a batch size of 32 on \gqa with the same AdamW optimizer and learning rate. \lxmt takes a batch size of 32 on \gqa and \vqa with the Adam optimizer and a learning rate of 1e-5.

\begin{table*}[ht]
    \centering
    \resizebox{1\linewidth}{!}{
    \begin{tabular}{cccc}
        \toprule
        Original operation & Compositional operation & Additional arguments & Additional comments \\
        \midrule
        \textlogic{some} & \textlogic{map} & \textlogic{or} \\
        \textlogic{all} & \textlogic{map} & \textlogic{and} \\
        \textlogic{none} & \textlogic{map} & \textlogic{or} & add \textlogic{logic\_not} in front of it \\      
        \textlogic{choose\_name} & \textlogic{choose} & \textlogic{name} \\
        \textlogic{choose\_attr} & \textlogic{choose} & \textlogic{attr} \\
        \textlogic{choose\_relation} & \textlogic{choose} & \textlogic{rel} \\
        \textlogic{query\_name} & \textlogic{query} & \textlogic{name} \\
        \textlogic{query\_attr} & \textlogic{query} & \textlogic{attr} \\
        \textlogic{verify\_attr} & \textlogic{verify} & \textlogic{attr} \\
        \textlogic{with\_relation} & \textlogic{filter} & \textlogic{rel} \\
        \textlogic{with\_relation\_object} & \textlogic{filter} & \textlogic{rel}, \textlogic{other} \\
        \textlogic{filter} & \textlogic{filter} & \textlogic{attr} \\  
        \textlogic{relation\_between\_nouns} &  &  & merged with \textlogic{choose\_relation} \\  
        \textlogic{find} & \textlogic{find} \\  
        \textlogic{count} & \textlogic{count} \\  
        \textlogic{keys} & \textlogic{keys} \\  
        \textlogic{unique\_images} & \textlogic{unique\_images} \\  
        \textlogic{group\_by\_images} & \textlogic{group\_by\_images} \\  
        \textlogic{scene} & \textlogic{scene} \\  
        % \textlogic{all\_same} & \textlogic{all\_same} \\
        \textlogic{exists} & \textlogic{exists} \\  
        \textlogic{logic\_or} & \textlogic{logic\_or} \\  
        \textlogic{logic\_and} & \textlogic{logic\_and} \\
        \textlogic{} & \textlogic{logic\_not} &  & new operation \\
        \textlogic{keep\_if\_values\_count\_eq} & \textlogic{keep\_if\_values\_count} & \textlogic{eq} \\
        \textlogic{keep\_if\_values\_count\_geq} & \textlogic{keep\_if\_values\_count} & \textlogic{geq} \\
        \textlogic{keep\_if\_values\_count\_leq} & \textlogic{keep\_if\_values\_count} & \textlogic{leq} \\
        \textlogic{eq} & \textlogic{compare} & \textlogic{eq} \\
        \textlogic{geq} & \textlogic{compare} & \textlogic{geq} \\
        \textlogic{leq} & \textlogic{compare} & \textlogic{leq} \\
        \textlogic{lt} & \textlogic{compare} & \textlogic{lt} \\
        \textlogic{gt} & \textlogic{compare} & \textlogic{gt} \\
        \textlogic{unique} &  &  & replaced with grammar checker \\
        \textlogic{assert\_unique} &  &  & replaced with grammar checker \\

        \bottomrule
    \end{tabular}
    }
    \caption{\newlogicform Creation. Refine 33 original operations into 18 compositional operations with additional arguments.}
    \label{tab:clf_operations}
\end{table*}
\begin{table*}[ht]
    \centering
    \tabcolsep 10pt
    \resizebox{1\linewidth}{!}{
    \begin{tabular}{llccc}
        \toprule
        Template & Transfer Setting & K-shot & \multicolumn{1}{c}{\vlendtoend} & \multicolumn{1}{c}{\ns} \\ 
        \cmidrule(lr){4-4}\cmidrule(lr){5-5}
        & & & \visb & \newlogicform w/ \tfive \\
        \midrule
        \multirow{3}{*}{\countgb}
        & \multirow{3}{*}{\textnlp{at least} $\rightarrow$ \textnlp{no less than}}  & 0 & \bf 54.0 & 21.5 [39.4]\\
        & & 1 & \bf 53.3 & 28.8 [47.5] \\
        & & 5 & \bf 56.3 & 39.2 [70.3] \\
        \midrule
        \multirow{6}{*}{\vcount}
        & \multirow{3}{*}{\textnlp{at least} $\rightarrow$ \textnlp{no less than}} & 0 & \bf 68.3 & 48.4 [67.8] \\
        & & 1 & \bf 68.8 & 55.1 [77.9] \\
        & & 5 & \bf 78.7 & 69.8 [91.1] \\
        \cmidrule{3-5}
        & \multirow{3}{*}{\textnlp{at least} $\rightarrow$ \textnlp{less than}} & 0 & 21.1 & {\bf 52.9} [65.6] \\
        & & 1 & 20.1 & {\bf 58.0} [73.3] \\
        & & 5 & 45.5 & {\bf 70.7} [91.5] \\
        \midrule
        \multirow{6}{*}{\vcountgb} 
        & \multirow{3}{*}{At least $\rightarrow$ No less than} & 0 & \bf 65.8 & 55.6 [76.2] \\
        & & 1 & \bf 66.2 & 59.5 [82.3] \\
        & & 5 & \bf 68.3 &  65.2 [92.0] \\
        \cmidrule{3-5}
        & \multirow{3}{*}{\textnlp{at least} $\rightarrow$ \textnlp{less than}} & 0 & 28.5 & {\bf 51.1} [50.6] \\
        & & 1 & 29.4 & {\bf 58.0} [70.5] \\
        & & 5 & 51.9 & {\bf 66.5} [86.9] \\
        \midrule
        \multirow{12}{*}{\quant} & \multirow{3}{*}{\textnlp{no} $\leftrightarrow$ \textnlp{some}} & 0 & 74.9 & {\bf 75.6} [97.8]  \\
        & & 1 & 73.5 & {\bf 76.1} [99.5] \\
        & & 5 & 75.4 & {\bf 76.1} [99.5] \\
        \cmidrule{3-5}
        & \multirow{3}{*}{\textnlp{no} $\rightarrow$ \textnlp{at least one}} & 0 & \bf 83.9 & 53.9 [61.1] \\
        & & 1 & \bf 84.5 & 61.2 [80.2] \\
        & & 5 & \bf 84.6 & 71.3 [95.1] \\
        \cmidrule{3-5}
        & \multirow{3}{*}{\textnlp{some} $\rightarrow$ \textnlp{none of the}} & 0 & 67.8 & {\bf 75.5} [97.8] \\
        & & 1 & 67.3 & {\bf 75.8} [98.6] \\
        & & 5 & 66.3 & {\bf 75.9} [99.5] \\
        \cmidrule{3-5}
        & \multirow{3}{*}{\textnlp{all} $\rightarrow$ \textnlp{either none or only some}} & 0 & 36.2 & {\bf 57.5} [69.8] \\
        & & 1 & 52.5 & {\bf 69.2} [89.0] \\
        & & 5 & 56.0 & {\bf 73.9} [97.9] \\
        \bottomrule
    \end{tabular}
    }
    \caption{Few-shot Contrast Set on \covr: Comparison of \visb and the \ns method on language contrast sets of four templates on \covr. \ns performance quickly improve from 1-shot training, \vlendtoend instead, needs more examples to learn, except for 1 unnatural perturbation, \textnlp{all}$\rightarrow$\textnlp{either none or only some}.}
    \label{tab:few-shot-contrast_set-covr}
\end{table*}
\subsection{Few-shot Training}
\label{suppsubFewshot}
Corresponding to Figure~\ref{fig:few_shot}, the whole few-shot training table is Table~\ref{tab:few-shot-contrast_set-covr}.

% \subsection{Analysis of the Exact Match Score}
% \label{suppsubExactM}

% \section{Full Related Works}
% \label{suppRelated}
% 
% 1. compositional generalization
% 2. contrast set?
% 3. scene graphs

% \section{\bill{Anything missing?}}

\end{document}